\documentclass[letterpaper]{article}

\usepackage{aaai2027}
\nocopyright
\usepackage[hyphens]{url}
\usepackage{graphicx}
\usepackage{natbib}
\usepackage{caption}
\usepackage{booktabs}
\usepackage{amsmath}
\usepackage{amssymb}
\usepackage{tabularx}
\usepackage{array}

\newcolumntype{Y}{>{\raggedright\arraybackslash}X}
\newcolumntype{C}{>{\centering\arraybackslash}X}
\newcommand{\pp}{\,\mathrm{pp}}

\newcommand{\NA}{--}
\newcommand{\R}{\textsc{Reject}}
\newcommand{\V}{\textsc{Validate}}
\newcommand{\T}{\textsc{Train}}

\title{Knowing When Not to Reuse: Conditional Experience Transfer in Autonomous LLM Post-Training}

\author{
Tingyun Li, Wenfeng Feng, Weiqing Li,\\
Abudukelimu Wuerkaixi, Guohua Liu, Yuewei Zhang
}
\affiliations{
Alibaba Cloud Computing\\
\{litingyun.lty,wenfeng.fwf,liyou.zyw\}@alibaba-inc.com\\
suoni@taobao.com
}

\begin{document}

\maketitle

\begin{abstract}
Large language models offer broad capabilities, but adapting them to evolving domains, tools, and requirements often entails repeated post-training. Autonomous systems automate parts of this process by proposing updates, training candidates, and using evaluation feedback to select subsequent proposals. As evidence accumulates, a central problem emerges: which past update evidence remains actionable after subsequent training has changed the parent model? An update's effect depends on its parent, data, and training stage. Treating past success as context-free permission can waste compute. If the resulting child is promoted, it can also degrade the subsequent training trajectory. We formulate this problem as \emph{conditional experience transfer} and introduce Boundary-Calibrated Intervention Transfer (BCIT), a method that authorizes experience reuse before weight-changing training. BCIT binds an observed effect to its source context, checks applicability conditions, vetoes candidates with named hard conflicts, and obtains current-state evidence through a bounded training trial when needed. Fully trained candidates still face a shared adoption rule, and only observed events extend memory. On one 4B model adapted across finance reasoning, text-to-SQL, and function calling, candidate updates exhibit heterogeneous target and retention effects across the evaluated contexts. Under matched candidates, evidence, and compute, BCIT authorizes fewer harmful updates and attains higher equal-budget final-model quality than the evaluated alternatives. These results support treating experience authorization as a distinct problem in autonomous post-training.
\end{abstract}

\section{Introduction}

Large language models can solve a broad range of tasks, but adapting them to new domains, tools, and requirements often demands repeated post-training. Autonomous systems automate parts of this loop: they propose an update, train a candidate, evaluate it, and use the feedback to select or revise subsequent proposals \citep{Yano2025LaMDAgent,Rank2026PostTrainBench,Ma2026TREX,Chen2026EvoTrainer,Chen2026Agent2RLBench}. The loop also accumulates a potentially valuable history of updates and outcomes.

That history is not self-executing. An update's effect depends on its parent model, data mixture, training stage, and evaluation contract. Evidence that an update was beneficial under one source context may therefore be misleading under the current context. Incorrectly authorizing a context-incompatible update consumes scarce training budget; if the trained child is then promoted, it also changes the parent checkpoint and the relevance of later evidence. The resulting problem is broader than storage or retrieval: \emph{how can an autonomous system use past updates without turning context-bound evidence into context-free instructions?} We call this problem \emph{conditional experience transfer}.

This framing follows a basic lesson from learning-transfer research: transfer depends on both what is reused and the conditions of reuse \citep{Barnett2002FarTransfer}. Toulmin's model of practical argument offers a complementary analogy: evidence supports a claim through an applicable warrant, while explicit exceptions can defeat it \citep{Toulmin2003Argument}. Here, an observed gain is evidence that an update was beneficial under its source model and conditions; it is not unconditional permission to spend full-training compute from a changed parent. Prior work searches for training configurations, organizes experience, estimates task- or model-level transferability, or allocates partial budgets. It does not resolve the candidate-level decision studied here: whether source-context evidence justifies rejecting, validating, or fully training a concrete candidate from the current parent \citep{Aamodt1994CaseBasedReasoning,Zhao2024ExpeL,Wang2025AWM,Nguyen2020LEEP,You2021LogME,Li2018Hyperband,Li2020ASHA,Chwa2026AutoPipe,Chen2026EvoTrainer}.

We introduce \textbf{Boundary-Calibrated Intervention Transfer (BCIT)} to make this decision explicit. BCIT records each update with the source context in which its effect was observed, the strength and provenance of that evidence, prespecified applicability conditions, and named hard conflicts. For the current parent, BCIT takes one of three actions. A non-compensable conflict causes rejection, unresolved evidence triggers a bounded current-parent trial, and satisfying the frozen rule authorizes full training. Authorization does not imply promotion: every fully trained child faces the same adoption rule before it can replace the parent. When the frozen library lacks coverage or progress stalls, an agent may generate up to three new atomic proposals. These proposals have no observed source effect, so none can reach full training without passing validation. Only executed events extend memory.

We evaluate this decision process while adapting one Qwen3-4B model for finance reasoning, text-to-SQL, and function calling, with instruction following as a retention constraint. Sequential promotions move the shared model through a series of parent checkpoints, creating a stress test for experience transfer; the three capabilities are the evaluation setting, not the method's objective. The experiments test four linked claims. Update effects vary across evaluated contexts. Under matched information, BCIT authorizes fewer harmful candidates while retaining beneficial ones. Bounded current-parent validation provides useful but imperfect evidence. The complete policy improves final-model quality under equal compute.

Our contributions are:
\begin{itemize}
    \item We formulate conditional experience transfer as a distinct control problem in autonomous post-training: because each promoted child becomes the parent for later decisions, evidence from past updates must remain bound to its source conditions.
    \item We introduce BCIT, a transparent reject--validate--train method that combines source evidence, explicit applicability conditions, non-compensable conflicts, bounded proposal generation, and a shared promote-or-rollback rule.
    \item We evaluate the full evidence chain through matched candidate--context diagnostics, outcome-blind authorization audits, short-to-full validation fidelity, and paired equal-budget episodes against validation-intensive and same-evidence alternatives.
\end{itemize}

\section{Related Work}

\paragraph{Autonomous Post-Training.}
Agent-driven systems automate data selection, training, evaluation, and iterative pipeline revision. LaMDAgent and TREX construct or revise post-training pipelines, while PostTrainBench and Agent$^2$ RL-Bench evaluate such agents under controlled compute \citep{Yano2025LaMDAgent,Ma2026TREX,Rank2026PostTrainBench,Chen2026Agent2RLBench}. AutoML systems likewise use prior runs: AutoPipe learns dataset-conditioned configuration rankings, and EvoTrainer tracks model versions, diagnostics, failures, and reusable training skills \citep{Feurer2015AutoSklearn,Chwa2026AutoPipe,Chen2026EvoTrainer}.These systems improve candidate generation and iterative training. BCIT studies a complementary pre-full-training decision: whether source-context evidence warrants rejecting a concrete candidate, validating it from the current parent, or allocating its full-training budget after the parent has changed.

\paragraph{Experience Transfer.}
Taskonomy maps task relations, while LEEP and LogME estimate transferability at the representation or pretrained-model level \citep{Zamir2018Taskonomy,Nguyen2020LEEP,You2021LogME}. Negative-transfer studies ask when source information harms a target \citep{Wang2019NegativeTransfer}. Case-based reasoning and agent-memory methods retrieve and revise past cases, reflections, or workflows \citep{Aamodt1994CaseBasedReasoning,Shinn2023Reflexion,Zhao2024ExpeL,Wang2025AWM}. These approaches typically estimate transferability at the task or model level, or retrieve reusable content. To our knowledge, they do not evaluate the same decision studied here: whether the observed source-context effect of one weight-changing candidate is sufficient to allocate validation or full-training budget from a changed parent. BCIT represents that decision using source evidence, explicit applicability conditions, and non-compensable conflicts.

\paragraph{Sequential Model Adaptation.}
Task arithmetic applies fine-tuning deltas, while model soups and TIES-Merging combine trained models or parameters and address compatibility at integration time \citep{Ilharco2023TaskArithmetic,Wortsman2022ModelSoups,Yadav2023TIES}. Continual learning studies how to acquire capabilities while limiting forgetting across sequential updates \citep{LopezPaz2017GEM}. BCIT acts before full training or model integration: it determines whether source-bound evidence warrants compute from the current parent. If a candidate is fully trained, ordinary post-training, continual-learning, or integration procedures may then handle the child under the shared adoption rule.

\section{Problem Formulation}

\subsection{State-Bound Decision Unit}

At decision step $t$, the current training context is
\begin{equation}
x_t=(m_t,D_t,\mathcal P_t,\ell_t),
\end{equation}
where $m_t$ is the parent model, $D_t$ identifies the training data and provenance available at this step, $\mathcal P_t$ is the evaluation and output protocol, and $\ell_t$ records the training stage and retention requirements. A candidate specifies an \emph{atomic update} $i$. When executed from a fixed parent, $i$ changes one declared data or training factor through a reproducible configuration difference. Together with the matched-control design used for paired contrasts, this restriction makes the effect of the declared intervention interpretable.

A historical candidate is represented by the state-bound record
\begin{equation}
c_i^{\mathrm{hist}}=(i,x_s,\mathcal E_s,\mathcal B_i,\rho_i),
\end{equation}
where $x_s$ is the source context, $\mathcal E_s$ is the observed source evidence, $\mathcal B_i$ contains prespecified applicability conditions and named conflicts, and $\rho_i$ links the model, data, configuration, and evaluation artifacts. The record supports only the effect observed at $x_s$; it does not assign a context-free positive or negative label to $i$. A new proposal generated under the fixed exploration quota has the corresponding form $c_i^{\mathrm{prop}}=(i,\bot,\bot,\mathcal B_i,\rho_i)$: its update, conditions, and provenance are specified, but no source context or outcome is fabricated. We write $c_i$ when the distinction is not needed.

The decision unit is the pair $(c_i,x_t)$: one candidate considered for one current context. If the candidate receives full training budget,
\begin{align}
m_t^+(i)&=\operatorname{Train}(m_t,i,b_i^{\mathrm{full}}),\\
\boldsymbol\Delta^{\mathrm{out}}(i;x_t)
&=M_{\mathrm{out}}(m_t^+(i))-M_{\mathrm{out}}(m_t),
\end{align}
where $M_{\mathrm{out}}$ is the prespecified vector of target and retention metrics used to score the full-run outcome. Let $\kappa_{\mathrm{tar}}>0$ be the minimum target improvement, $\epsilon_r\geq0$ the permitted loss on retention metric $r$, and $F_i^{\mathrm{out}}\in\{0,1\}$ indicate a hard execution failure. The context-indexed outcome is Beneficial when $\Delta_{\mathrm{tar}}^{\mathrm{out}}\geq\kappa_{\mathrm{tar}}$, every $\Delta_r^{\mathrm{out}}\geq-\epsilon_r$, and $F_i^{\mathrm{out}}=0$. It is Harmful when the target degrades, any retention bound is violated, or $F_i^{\mathrm{out}}=1$. All other outcomes are Neutral. This label exists only after a full outcome is observed. It is used for retrospective scoring and is distinct from the adoption decision below.

\subsection{Authorization, Adoption, and Budget}

Before full training, an authorization policy chooses
\begin{equation}
a(c_i,x_t)\in\{\textsc{Reject},\textsc{Validate},\textsc{Train}\}.
\end{equation}
\textsc{Reject} declines the candidate in the current context. \textsc{Validate} runs a real, budget-capped training trial from $m_t$ to obtain current-state evidence; its temporary checkpoint is never promoted. \textsc{Train} allocates the budget required to execute the full update from $m_t$. Retrieval or proposal generation alone can neither allocate this budget nor replace the current parent.

After full training, a separate rule $g$, fixed across the compared authorization policies, determines the next parent:
\begin{equation}
m_{t+1}=\begin{cases}
m_t^+(i), & g(m_t,m_t^+(i))=\textsc{Promote},\\
m_t, & g(m_t,m_t^+(i))=\textsc{Rollback}.
\end{cases}
\end{equation}
Thus, historical evidence or a bounded validation can allocate full-training compute, but neither can replace the parent directly. This separation also keeps two empirical questions distinct: whether authorization selected a beneficial full run, and whether the shared adoption rule promoted the trained child.

Given total budget $B_{\mathrm{gpu}}$, a sequential policy $\pi$ seeks a high-utility final model while satisfying retention constraints:
\begin{align}
\max_{\pi}\quad &U(m_{\pi}(B_{\mathrm{gpu}}))\\
\text{s.t.}\quad &C_{\pi}\leq B_{\mathrm{gpu}},
\qquad G_r(m_{\pi}(B_{\mathrm{gpu}}))\geq-\epsilon_r\quad\forall r.
\end{align}
Here, $U$ is the prespecified final-model utility, $C_{\pi}$ charges validation, full training, evaluation, loading, and failed runs, and $G_r$ is retention change relative to the initial parent. Before each evaluation episode, the initial experience library, candidate generator, data roles, decision thresholds, validation fallback, evaluators, adoption rule, and budget are frozen. Final test outcomes remain sealed until the episode terminates.

\section{Boundary-Calibrated Intervention Transfer}

\begin{figure*}[t]
  \centering
  \includegraphics[width=0.99\textwidth]{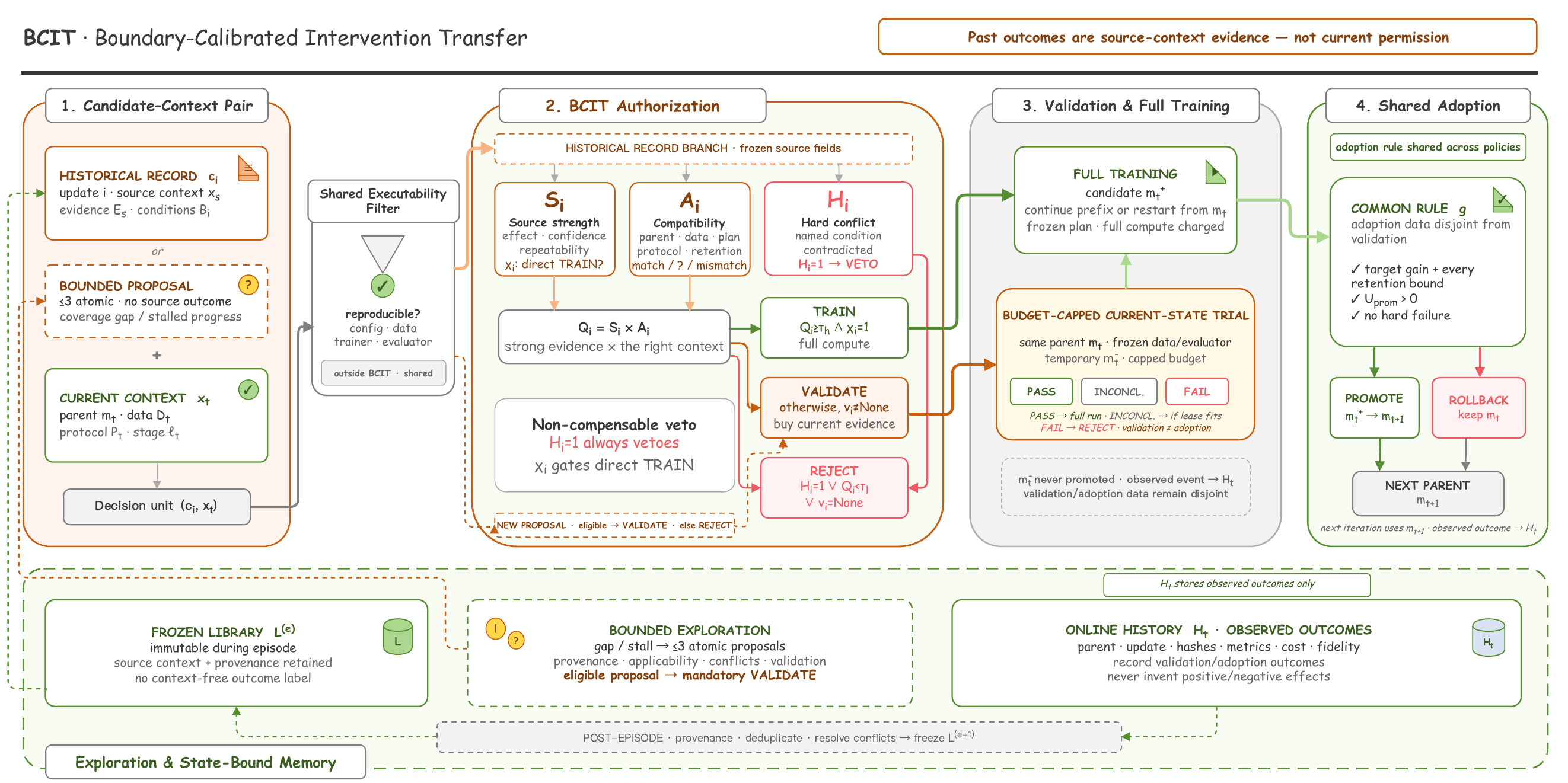}
  \caption{BCIT uses source-context evidence to authorize candidates for full training. Historical records combine source strength, current compatibility, and a hard-conflict veto. A new proposal has no source score: it enters validation only if it passes the shared executability check, has no hard conflict, and admits a faithful short run; otherwise it is rejected. A validation checkpoint is never promoted. Every fully trained child faces the same adoption rule, and only observed events extend memory.}
  \label{fig:bcit-overview}
\end{figure*}

BCIT sits between candidate generation and full, weight-changing training (Figure~\ref{fig:bcit-overview}). After a shared executability check, it evaluates a historical candidate using source strength $S_i$, current compatibility $A_i$, and a non-compensable hard-conflict indicator $H_i$. It then chooses \textsc{Reject}, bounded \textsc{Validate}, or full \textsc{Train}. A new proposal has no observed source effect and cannot bypass validation. Validation checkpoints are temporary; a rule shared across policies promotes or rolls back every fully trained child.

% Standalone float module; include from the project root.
% Source label: fig:bcit-overview
% Float kind: figure

\subsection{Evidence and Applicability}

All policies first discard candidates whose configuration, data, trainer, or evaluator cannot be reproduced. The same filter precedes every compared policy, so its exclusions are not attributed to BCIT.

For an executable historical candidate, $\delta_i^{\mathrm{src}}$ is the frozen signed change in the source task's prespecified primary metric, expressed in raw score units. We use the lower confidence bound when repeated paired runs support an interval. Otherwise, we use the recorded point estimate. Its \emph{source strength} is
\begin{equation}
S_i=d(e_i)\operatorname{clip}
\left(\frac{\delta_i^{\mathrm{src}}}{\kappa_i},0,1\right),
\end{equation}
where $\kappa_i$ is the prespecified normalization scale, $e_i$ is the frozen evidence grade, and $d(e_i)\in(0,1]$ is its discount. Grades allowed to skip validation form $\mathcal E_{\mathrm{direct}}$, with $\chi_i=\mathbf{1}[e_i\in\mathcal E_{\mathrm{direct}}]$. In our protocol, $d(A/B/C)=1/.75/.5$ and $\mathcal E_{\mathrm{direct}}=\{A\}$; only replicated matched-control evidence receives grade A. A point estimate may affect $S_i$ but cannot bypass validation.

\emph{Current compatibility} averages the update's prespecified applicability conditions:
\begin{equation}
A_i=\frac{1}{J_i}\sum_{j=1}^{J_i}a_{ij},
\qquad a_{ij}\in\{0,0.5,1\},
\end{equation}
where the values denote mismatch, unresolved, and match. Conditions use only information fixed before the candidate is executed, including the parent stage, data composition, optimization regime, output interface, and evaluation contract. \emph{Hard conflict} $H_i=1$ only when a named required condition is contradicted; missing information is treated as unresolved.

These human-specified fields are frozen before target outcomes open. A deterministic \emph{boundary compiler} reads only source records and structured current-context metadata and cannot read target outcomes; timestamps and configuration hashes prevent retrospective revision. BCIT evaluates authorization under these fixed boundaries rather than learning them.

BCIT combines the two positive signals as
\begin{equation}
Q_i=S_iA_i.
\end{equation}
Consequently, high source strength cannot compensate for low current-context compatibility. The product is a prespecified transparent score, not a learned optimum.

\subsection{Authorization}

The compiler also assigns $\nu_i\in\{\textsc{Continue},\textsc{Restart},\textsc{None}\}$: validation is a prefix of the full run, a separate short proxy followed by restart, or unavailable because no faithful short run exists. An executable historical candidate follows
\begin{equation}
a(c_i^{\mathrm{hist}},x_t)=
\begin{cases}
\textsc{Reject}, & H_i=1\ \lor\ Q_i<\tau_l,\\
\textsc{Train}, & Q_i\geq\tau_h\ \land\ \chi_i=1,\\
\textsc{Validate}, & \nu_i\neq\textsc{None},\\
\textsc{Reject}, & \nu_i=\textsc{None}.
\end{cases}
\end{equation}
The thresholds are calibrated on units disjoint from the primary audit and frozen before its outcomes open; we use $\tau_l=.30$ and $\tau_h=.70$. Direct training additionally requires $\chi_i=1$, preventing weak source evidence from bypassing validation.

A new proposal receives neither $S_i$ nor $\chi_i$: provenance establishes reproducibility, not effectiveness. Its route is
\begin{equation}
a(c_i^{\mathrm{prop}},x_t)=
\begin{cases}
\textsc{Validate}, & H_i=0\ \land\ \nu_i\neq\textsc{None},\\
\textsc{Reject}, & \text{otherwise}.
\end{cases}
\end{equation}
Thus, a proposal can receive full-training budget only through a faithful current-state validation, never through an artificial source score.

\subsection{Current-State Validation}

Validation runs the candidate update from the current parent under a capped budget:
\begin{align}
\widetilde m_t&=\operatorname{Train}(m_t,i,b_i^{\mathrm{val}}),
\qquad b_i^{\mathrm{val}}\leq\bar b^{\mathrm{val}},\\
\boldsymbol\Delta_i^{\mathrm{val}}&=
M_{\mathrm{val}}(\widetilde m_t)-M_{\mathrm{val}}(m_t).
\end{align}
The short run starts from $m_t$; its change is measured against the unmodified parent using the same frozen evaluation data, evaluator, and metrics. The result is \textsc{Pass} when the target reaches its validation threshold, all retention bounds hold, and no format, parsing, or execution failure occurs. It is \textsc{Fail} when a frozen degradation or guardrail condition fires. All other results are \textsc{Inconclusive}.

A \textsc{Pass} routes the candidate to full training. A \textsc{Fail} rejects it in the current context, and the temporary checkpoint is never promoted. An \textsc{Inconclusive} result routes the candidate to full training only when the frozen fallback reserves a full-run lease and the remaining budget can pay for it; otherwise the parent remains unchanged. This fallback is a study setting, not a consequence of $Q_i$.

For \textsc{Restart}, both the proxy and restarted full run are charged. The fidelity study evaluates every compiled short/full pair rather than only passed trials.

\subsection{Shared Adoption}

Every fully trained candidate is evaluated by fixed rule $g$ on adoption data disjoint from validation. The rule promotes the child only if the target criterion and all retention constraints are satisfied and no hard failure occurs. Otherwise, it retains the parent. This utility and promote-or-rollback rule are fixed and shared across all compared authorization policies.

\subsection{Memory and Exploration}

The curated library $\mathcal L^{(e)}$ is immutable within episode $e$. A separate history $\mathcal H_t$ records each decision's parent, update, hashes, action, and cost; executed validations and full runs also record metrics, fidelity, and outcomes. Rejections record only their frozen reason and create no effect label, while observed failures remain bound to their parent. Online events may be retrieved but never overwrite their source records.

After the episode, a new library version can be created only through
\begin{equation}
\mathcal L^{(e+1)}=\operatorname{Consolidate}
\bigl(\mathcal L^{(e)},\mathcal H_T^{(e)}\bigr),
\end{equation}
which applies provenance checks, deduplication, and conflict review before the next freeze. New evidence can therefore affect later episodes without rewriting an ongoing one.

When a prespecified coverage gap or stall occurs, the agent may consult a frozen, provenance-audited knowledge snapshot and generate at most $q$ atomic proposals. We set $q=3$. Each proposal specifies one primary change, its configuration difference, expected effect, conditions, conflicts, and validation plan. The proposal-generating agent cannot alter evaluators or rewards, access sealed outcomes, or bundle uncontrolled interventions. Every proposal re-enters the common checks and must validate.

\subsection{Controlled Authorization Comparators}

Flat-Additive receives the same candidates, evidence, applicability fields, proposal route, validation, executor, adoption rule, and budget. For historical candidates, it replaces BCIT's product and hard-conflict veto with
\begin{equation}
Z_i^{\mathrm{flat}}=\{S_i+A_i+(1-H_i)\}/3,
\end{equation}
so source evidence can compensate for poor context match and a conflict becomes an ordinary feature. Additive+Veto retains BCIT's hard rejection and all downstream rules, changing only the positive combination to $Z_i^{\mathrm{add}}=(S_i+A_i)/2$. Validate-All ignores $S_i$, $A_i$, and $H_i$ after the common executability check and runs bounded validation for every executable candidate. Each score-based policy calibrates its own thresholds because the score distributions differ. BCIT without hard veto removes only the $H_i=1$ rejection. BCIT-Reject-Unresolved removes validation by rejecting every candidate routed to \textsc{Validate}; direct-training and adoption rules are unchanged. Because authorization changes later parents and opportunities, end-to-end contrasts compare complete equal-budget policies; the three-seed component runs remain descriptive.

\section{Experiments and Results}

\subsection{Questions and Protocol}

The experiments test four linked claims. RQ1 measures effect heterogeneity across matched candidate--context pairs. RQ2 compares authorization policies given identical evidence. RQ3 tests whether bounded current-parent validation predicts full training. RQ4 compares complete policies under matched compute.

All runs start from Qwen3-4B. Experience from FinQA, Spider, and xLAM is transferred to TAT-QA, BIRD, and BFCL, with IFEval as a common retention constraint \citep{Chen2021FinQA,Yu2018Spider,Zhang2024xLAM,Zhu2021TATQA,Li2023BIRD,BFCL2024,Zhou2023IFEval}. Sequential promotions move the shared model through changing parent checkpoints, providing a stress test for experience transfer. The primary endpoint is
\begin{equation}
M_{\mathrm{avg}}=\tfrac{1}{3}(M_{\mathrm{TATQA}}+M_{\mathrm{BIRD}}+M_{\mathrm{BFCL}}),
\end{equation}
with prompt- and instruction-level IFEval guardrails at a prespecified $-2$-point margin relative to Base. On adoption data disjoint from validation, the shared promotion utility is
\begin{equation}
U_{\mathrm{prom}}=\frac{1}{|\mathcal F|}
\sum_{r\in\mathcal F}\frac{\Delta_r^{\mathrm{prom}}}{\kappa_r},
\end{equation}
where $\mathcal F$ is the capability set and $\kappa_r$ is the minimum meaningful change for capability $r$. Promotion requires the target gain, all retention bounds, $U_{\mathrm{prom}}>0$, and no hard failure. We also report worst task gain and the budget-normalized area under $M_{\mathrm{avg}}$ versus consumed GPU-hours (AUC).

Audit-24 contains 10 beneficial, 8 harmful, and 6 neutral outcomes. Policies see identical inputs, freeze decisions before full outcomes open, and are then scored on those outcomes. BCIT, Flat-Additive, Validate-All, and Additive+Veto use six paired end-to-end seeds with the same start, candidate stream, data order, adoption rule, and 36-GPU-hour cap. BCIT-Retrieve, the two component ablations, and the remaining shared-model baselines use three seeds descriptively; each task specialist uses one run. Validate-All and Additive+Veto were added after the original BCIT--Flat study. Before either new baseline outcome was opened, the amendment prespecified BCIT--Validate-All as the first comparison and BCIT--Additive+Veto as the second. We report paired differences, paired $t$ intervals, and exact sign-flip tests. Exact interval definitions, exchangeability assumptions, baseline contracts, and seed-level results are in the supplement.

\subsection{How Stable Are Update Effects Across Evaluated Contexts? (RQ1)}

We retrospectively audit 24 matched candidate--context pairs, eight per capability. Each control/intervention pair shares its parent, seed, data scale, evaluation examples, and budget. Because outcomes were exposed before analysis, this cohort diagnoses heterogeneity but is not used to claim policy superiority.

% Standalone float module; include from the project root.
% Source label: fig:heterogeneity
% Float kind: figure
\begin{figure*}[t]
    \centering
    \includegraphics[width=0.98\textwidth]{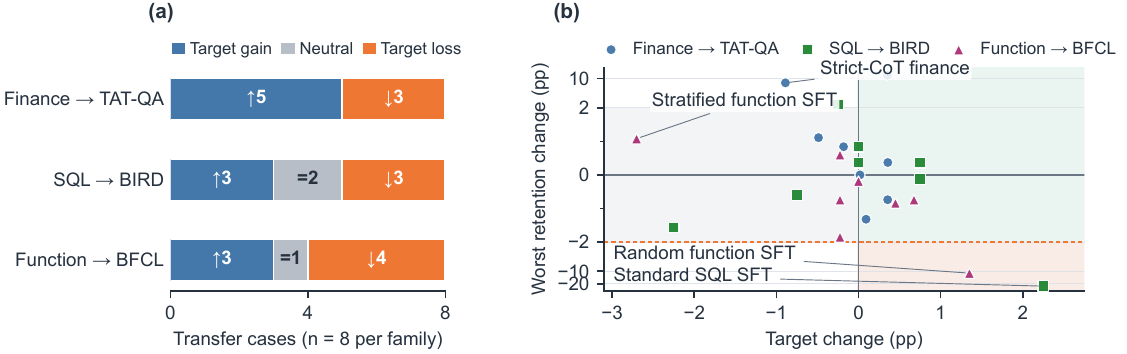}
    \caption{Candidate updates exhibit heterogeneous target and retention effects across 24 matched candidate--context pairs. (a) Target-direction counts ($n=8$ per family); arrows redundantly encode direction. (b) Target change versus the worse IFEval change. The gray $[-2,2]$ band is linear; the disclosed symmetric-log scale compresses only larger retention changes. Semantic labels replace run identifiers.}
    \label{fig:heterogeneity}
\end{figure*}

Figure~\ref{fig:heterogeneity} shows that 13 of 24 candidates do not improve their target. Only 3 of the 11 target-improving candidates also improve both retention measures. Two cases illustrate why target gain alone is insufficient: a standard supervised SQL update gains 2.25 target points but loses 22.74/18.71 IFEval-P/I points, while a random-sampling function-calling update gains 1.35 points but loses 11.09/8.15. The decision problem is therefore whether source-context evidence generalizes to the current parent and evaluation context, not merely whether an update has worked before. Candidate-level values are in the supplement.

\subsection{Does BCIT Reduce Harmful Authorization while Retaining Beneficial Candidates? (RQ2)}

We jointly report harmful-candidate authorization and beneficial-candidate coverage after decisions freeze. The former should be low and the latter high, preventing a trivial reject-all policy from being favored. Additional selective-decision metrics and marginal intervals are in the supplement.

% Standalone float module; include from the project root.
% Source label: tab:audit24
% Float kind: table
\begin{table*}[t]
\centering
\small
\begin{tabular}{@{}lrrrrr@{}}
\toprule
Policy & Authorized B/H/N & All auth. & Harmful auth. $\downarrow$ & Beneficial kept $\uparrow$ & Harmful share $\downarrow$ \\
\midrule
Reuse all & 10/8/6 & 100.0 & 100.0 & 100.0 & 33.3 \\
Source evidence only & 9/6/4 & 79.2 & 75.0 & 90.0 & 31.6 \\
Applicability only & 8/5/4 & 70.8 & 62.5 & 80.0 & 29.4 \\
Flat-Additive & 8/5/3 & 66.7 & 62.5 & 80.0 & 31.3 \\
\textbf{BCIT} & \textbf{9/2/3} & \textbf{58.3} & \textbf{25.0} & \textbf{90.0} & \textbf{14.3} \\
\midrule
BCIT w/o candidate-specific applicability & 9/6/3 & 75.0 & 75.0 & 90.0 & 33.3 \\
Additive+Veto & 8/3/3 & 58.3 & 37.5 & 80.0 & 21.4 \\
BCIT w/o hard veto & 9/4/3 & 66.7 & 50.0 & 90.0 & 25.0 \\
\bottomrule
\end{tabular}
\caption{Outcome-blind Audit-24. Policies receive the same frozen candidate records and current-context metadata, fix decisions before full outcomes open, and are scored against the same 10 beneficial, 8 harmful, and 6 neutral outcomes. Rates are percentages.}
\label{tab:audit24}
\end{table*}

BCIT authorizes 2 of 8 harmful candidates and 9 of 10 beneficial candidates, versus 5 and 8 for Flat-Additive (Table~\ref{tab:audit24}). Harmful authorization is therefore 25.0\% versus 62.5\%, while beneficial-candidate coverage is 90.0\% versus 80.0\%; among authorized candidates, the harmful share falls from 31.3\% to 14.3\%. The exact McNemar test gives $p=.25$, so this is directional cohort evidence rather than a population error rate. Removing candidate-specific applicability conditions raises harmful authorization to 75.0\%. Adding a veto to the additive score lowers it to 37.5\%, whereas removing the veto from BCIT raises it to 50.0\%. Within this cohort, the directions are consistent with complementary contributions from applicability conditions, multiplicative combination, and the hard veto.

\subsection{Does Bounded Current-Parent Validation Help? (RQ3)}

Every frozen candidate is executed once under the nominal 20\% validation budget and once under its matched full-training budget, regardless of the validation result. Short and full target directions agree for 20 of 24 candidates (83.3\%) with Spearman $\rho=.72$. The three-way screen returns 10 Pass, 9 Fail, and 5 Inconclusive decisions. Pass identifies Beneficial full-run outcomes with 80.0\% precision and 80.0\% recall at a median 17\% of full-training cost. Agreement rises from 62.5\% at a nominal 5\% budget to 75.0\% at 10\% and 83.3\% at 20\%, consistent with the 20\% primary operating point. Four sign reversals and two false Pass cases remain: a bounded-validation result is current-parent evidence, not a substitute for full training or shared adoption. The candidate-level plot, exact sign-agreement interval, and full ledger are in the supplement.

\subsection{Does the Complete Policy Improve Equal-Budget Outcomes? (RQ4)}

% Standalone float module; include from the project root.
% Source label: tab:system36
% Float kind: table
\begin{table*}[t]
\centering
\small
\begingroup
\setlength{\tabcolsep}{1mm}
\begin{tabular*}{0.80\textwidth}{@{\extracolsep{\fill}}lrrrrrcrr@{}}
\toprule
Method & TAT & BIRD & BFCL & Mean & Min. $\Delta$ & IFEval P/I & GPUh & AUC \\
\midrule
Base & 31.3 & 39.0 & 57.2 & 42.5 & 0.0 & 76.5/82.7 & 0.0 & 42.5 \\
\midrule
Fin. specialist & \textit{39.6} & -- & -- & -- & -- & -- & 36.0 & -- \\
SQL specialist & -- & \textit{49.3} & -- & -- & -- & -- & 36.0 & -- \\
Func. specialist & -- & -- & \textit{65.1} & -- & -- & -- & 36.0 & -- \\
\midrule
Flat-Add. & 33.4 & 41.2 & 58.5 & $44.4\pm0.3$ & +1.3 & 75.0/80.9 & $35.9\pm0.1$ & 43.6 \\
Validate-All & 34.6 & 42.5 & 59.3 & $45.5\pm0.2$ & +2.1 & 75.6/81.7 & $35.8\pm0.1$ & 44.2 \\
Add. + veto & 35.0 & 43.3 & 59.9 & $46.1\pm0.2$ & +2.7 & 76.0/82.1 & $35.3\pm0.2$ & 44.5 \\
No-memory & 33.2 & 40.8 & 57.8 & $43.9\pm0.1$ & +0.6 & 75.3/81.2 & $35.6\pm0.1$ & 43.5 \\
Source-only & 33.0 & 40.3 & 57.2 & $43.5\pm0.2$ & 0.0 & 74.3/80.4 & $35.3\pm0.1$ & 43.2 \\
Static multi-SFT & 32.3 & 39.6 & 56.1 & $42.6\pm0.2$ & $-1.1$ & 75.4/81.3 & $35.4\pm0.2$ & 42.8 \\
Reuse-all & 32.1 & 39.5 & 55.9 & $42.5\pm0.1$ & $-1.4$ & 73.6/79.3 & $35.5\pm0.2$ & 42.6 \\
\midrule
BCIT-Retrieve & 34.7 & 42.8 & 59.5 & $45.7\pm0.3$ & +2.3 & 76.0/82.0 & $34.7\pm0.2$ & 44.3 \\
BCIT w/o hard veto & 34.7 & 42.4 & 59.3 & $45.5\pm0.3$ & +2.1 & 75.2/81.3 & $35.6\pm0.1$ & 44.1 \\
BCIT-Reject-Unresolved & 34.4 & 42.3 & 58.9 & $45.2\pm0.2$ & +1.7 & 76.1/82.2 & $33.0\pm0.1$ & 43.9 \\
\textbf{BCIT} & \textbf{35.9} & \textbf{44.3} & \textbf{60.8} & $\mathbf{47.0\pm0.4}$ & \textbf{+3.6} & \textbf{76.7/82.9} & $35.1\pm0.3$ & \textbf{44.9} \\
\bottomrule
\end{tabular*}
\caption{36-GPU-hour endpoint results. TAT: TAT-QA; P/I: prompt-/instruction-level IFEval. Cross-task mean and GPUh are mean $\pm$ sample SD; other entries are means. Values are rounded to one decimal; tests use unrounded seed-level data. Bold marks the best shared-model value in each performance column.}
\label{tab:system36}
\endgroup
\end{table*}

% Standalone float module; include from the project root.
% Source label: fig:trajectory
% Float kind: figure
\begin{figure}[t]
    \centering
    \includegraphics[width=\columnwidth]{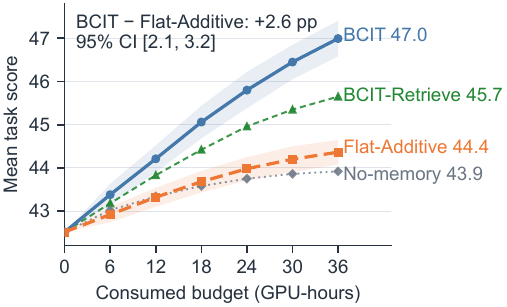}
    \caption{Observed equal-budget trajectories. Mean task score versus consumed budget; BCIT and Flat-Additive bands are $\pm1$ sample SD over six paired seeds.}
    \label{fig:trajectory}
\end{figure}

BCIT improves the final mean over Flat-Additive in all six paired runs (Table~\ref{tab:system36}). The mean gain is 2.63 points (95\% CI [2.10, 3.16]; exact sign-flip $p=.03125$), with gains of 2.48 on TAT-QA, 3.04 on BIRD, and 2.37 on BFCL. Mean IFEval scores remain above the prespecified margins. This supports the frozen guardrail in these episodes, not general non-inferiority; paired task and retention intervals are in the supplement.

The protocol amendment tests two stronger alternatives. Validate-All runs bounded validation for every executable candidate; Additive+Veto receives the same evidence and safeguards but combines positive signals additively. BCIT exceeds them by 1.50 points (95\% CI [0.85, 2.14]) and 0.90 points ([0.27, 1.52]), respectively. Each difference is positive in all six pairs ($p=.03125$), and BCIT uses 0.74 fewer GPU-hours than Validate-All. The endpoint advantage therefore persists against both prespecified alternatives and is not obtained by validating more candidates under this protocol.

The three-seed components are descriptive: removing the hard veto lowers the mean by 1.49 points; BCIT-Reject-Unresolved, which rejects every validation-routed candidate, lowers it by 1.76. These directions are consistent with both components contributing, but do not isolate component-level causal effects; the latter's early stop does not establish equal-quality efficiency. Specialists exceed BCIT on their named task but require three budgets. BCIT has the highest mean among shared models; BCIT-Retrieve's unpaired 45.7 is suggestive only.

The difference also appears before the endpoint (Figure~\ref{fig:trajectory}): BCIT's AUC is 44.9 versus 43.6 for Flat-Additive and 44.2 for Validate-All. The paired BCIT--Validate-All gain is 0.70 [0.22, 1.18]; the 0.47-point gap to Additive+Veto includes zero. Endpoint and trajectory evidence are therefore complementary, but AUC does not support superiority over every baseline. BCIT and Flat-Additive promotion counts, seed-level results for paired policies and component ablations, and the earlier 24-GPU-hour pilot appear in the supplement.

% \paragraph{Claim alignment across RQs.}
% BCIT rejects hard conflicts, validates unresolved evidence, and authorizes full training under a frozen rule.

\FloatBarrier

\section{Limitations}

Our evidence covers one 4B model, three target capabilities, one retention benchmark, and human-specified boundaries. The four studies share candidates and are complementary rather than independent replications; only complete-policy comparisons use six paired seeds, while components remain descriptive. Validation fidelity and compute trade-offs may vary with model scale, data, and schedule. Learned boundaries, broader domains, and longer trajectories remain future work.

\section{Conclusion}

Past success in one source context does not authorize modifying every future parent. BCIT enforces this distinction: it rejects named hard conflicts, validates unresolved candidates on the current parent, and grants full-training budget only under a frozen rule. A policy-shared adoption rule then promotes or rolls back each trained child. Across the evaluated evidence chain, update effects vary across contexts. Under matched information, BCIT authorizes fewer harmful candidates while retaining beneficial ones. Under matched compute, it attains a higher cross-task mean than all evaluated shared-model alternatives. These results support evidence-conditioned rather than unconditional allocation of post-training compute.

% Preserve the conference-paper reading order: main paper, references, then
% the complete supplementary material in the same PDF.
\clearpage
\bibliography{bcit_refs}

\clearpage
\appendix

% Supplement-only float policy.  This leaves the main-paper pagination and
% figure placement unchanged while keeping the appendix tables in order.
\setcounter{topnumber}{6}
\setcounter{bottomnumber}{6}
\setcounter{totalnumber}{12}
\setcounter{dbltopnumber}{8}
\renewcommand{\topfraction}{.99}
\renewcommand{\bottomfraction}{.99}
\renewcommand{\textfraction}{.01}
\renewcommand{\floatpagefraction}{.25}
\renewcommand{\dbltopfraction}{.99}
\renewcommand{\dblfloatpagefraction}{.20}
\makeatletter
\setlength{\@dblfptop}{0pt}
\setlength{\@dblfpsep}{4pt}
\setlength{\@dblfpbot}{0pt plus 1fil}
\makeatother

\twocolumn[
\begin{center}
\vspace*{0.12in}
{\LARGE\bfseries Supplementary Material\par}
\vspace{0.08in}
{\large\bfseries Knowing When Not to Reuse:\
Conditional Experience Transfer in Autonomous LLM Post-Training\par}
\vspace{0.20in}
\end{center}
]

\section{Scope and Study Separation}

This supplement provides additional controller specifications, experimental details, candidate- and seed-level results, and statistical analyses supporting the main paper. The four empirical components answer distinct questions and are kept separate throughout:

% Standalone float module; include from the project root.
% Source label: tab:study-map
% Float kind: table
\begin{table*}[!t]
\centering
\small
\caption{Relationship between the four empirical components. Unit rows are observed candidate-level or episode-level values rather than reconstructed pseudo-observations.}
\label{tab:study-map}
\begin{tabularx}{\textwidth}{@{}p{0.17\textwidth}p{0.22\textwidth}Y Y@{}}
\toprule
Study & Scientific role & Outcome access and sampling unit & Material reported here \\
\midrule
Retrospective-24 & Diagnose effect heterogeneity & Outcomes exposed before analysis; 24 matched candidate--context pairs & All target and retention deltas; semantic intervention labels \\
Audit-24 & Compare authorization decisions from matched information & Decisions frozen before full outcomes open; 24 decision units & Policy counts, structural ablations, exact intervals, and paired discordance table \\
ShortFull-24 & Measure bounded-validation fidelity & Every short/full pair executed; no pass-only filtering; 24 candidates & Full short/full ledger, sign reversals, confusion counts, and budget sensitivity \\
36-GPU-hour episodes & Compare complete sequential policies under equal compute & Six matched episodes for the primary and protocol-amendment comparisons & Seed-level endpoints, paired intervals, component directions, process counts, and aggregate endpoints \\
\bottomrule
\end{tabularx}
\end{table*}

Retrospective-24 documents heterogeneous effects across the evaluated candidate--context pairs but does not support a policy-superiority claim. Audit-24 evaluates authorization after decisions are frozen. ShortFull-24 evaluates a mechanism used by the policy but shares full outcomes with Audit-24. The end-to-end episodes evaluate complete sequential policies in which earlier decisions change later parent models and opportunities. No pooled effect is computed across these studies.

\section{Controller and Episode Contract}

Each end-to-end episode freezes
\begin{equation}
\Omega=(\mathcal{L}_0,\mathcal{K}_0,\mathcal{R},\Gamma,
D_{\mathrm{cal}},D_{\mathrm{val}},D_{\mathrm{prom}},D_{\mathrm{final}},g,\Theta,B),
\label{eq:omega}
\end{equation}
where the manifest contains the base-model identifier, seed, memory and knowledge snapshots, candidate inventory, retrieval order, evidence grades, minimum meaningful changes, score thresholds, validation compiler, proposal quota, evaluators, parsers, data roles, resource cap, stopping rule, and analysis specification. Final evaluation outcomes are sealed until the episode terminates and the final adopted checkpoint and GPU-hour ledger are fixed.

\subsection{Evidence Strength and Applicability}

The evidence-grade discount is
\begin{equation}
d(A)=1.00,\qquad d(B)=0.75,\qquad d(C)=0.50,
\end{equation}
where grade A denotes replicated matched-control evidence, grade B a single matched internal comparison, and grade C a provenance-checked external result or recipe. Only grade A may bypass current-state validation:
\begin{equation}
\mathcal{E}_{\mathrm{direct}}=\{A\},\qquad \chi_i=\mathbb{1}[e_i\in\mathcal{E}_{\mathrm{direct}}].
\end{equation}
Unscored anecdotes receive no evidence grade and cannot authorize training.

Here, $\delta_i^{\mathrm{src}}$ is the signed change in the source task's prespecified primary metric, expressed in raw score units. It is the lower confidence bound when replicated matched runs support an interval and the recorded point estimate otherwise. The normalized source strength follows the main paper,
\begin{equation}
S_i=d(e_i)\,\mathrm{clip}\!\left(\frac{\delta_i^{\mathrm{src}}}{\kappa_i},0,1\right).
\end{equation}
The controller uses a one-percentage-point task MCID and a source-score reference improvement of twice that MCID, so $\kappa_i=.02$ in raw score units. A negative source effect is clipped to zero rather than converted into positive evidence.

Applicability is the mean of five pre-training fields: capability-family match, declared training-data availability and identity, declared runtime availability, current-parent compatibility, and evaluator/output-protocol compatibility. Each field is scored $0$, $.5$, or $1$ for mismatch, unresolved, or match. A hard conflict $H_i=1$ requires a named contradiction in a required field; missing metadata remains unresolved and does not create a veto by itself. The positive score is
\begin{equation}
Q_i=S_iA_i,
\end{equation}
with frozen thresholds $\tau_l=.30$ and $\tau_h=.70$.

\subsection{Authorization and Validation Rule}

After a shared executability check, an executable historical candidate follows
\begin{equation}
 a(c_i^{\mathrm{hist}},x_t)=
 \begin{cases}
 \R, & H_i=1\ \text{or}\ Q_i<\tau_l,\\
 \T, & Q_i\ge \tau_h\ \text{and}\ \chi_i=1,\\
 \V, & \nu_i\ne \mathrm{None},\\
 \R, & \nu_i=\mathrm{None}.
 \end{cases}
\label{eq:authorize}
\end{equation}
A new proposal has neither $S_i$ nor $\chi_i$ and must validate when no hard conflict is present and a faithful short-run compilation exists. The exploration quota is at most three atomic proposals per episode. Proposals cannot alter evaluators, rewards, data roles, or sealed outcomes.

A validation is \textsc{Pass} when its target reaches the frozen validation threshold, every validation-time retention bound holds, and no format, parsing, or execution failure occurs. It is \textsc{Fail} when a prespecified target-degradation condition, retention violation, or hard failure fires, and \textsc{Inconclusive} otherwise. A \textsc{Pass} routes the candidate to full training. A \textsc{Fail} rejects it in the current context, and the temporary checkpoint is never promoted. An \textsc{Inconclusive} result routes the candidate to full training only when the frozen fallback reserved a full-run lease and the remaining budget can pay for it; otherwise the parent is unchanged. Every fully trained candidate then faces the same adoption rule $g$ on data disjoint from validation.

\paragraph{Shared adoption.} On adoption data disjoint from validation, the prespecified cross-capability utility is
\begin{equation}
U_{\mathrm{prom}}=\frac{1}{|\mathcal F|}
\sum_{r\in\mathcal F}\frac{\Delta_r^{\mathrm{prom}}}{\kappa_r},
\end{equation}
where $\mathcal F$ is the capability set and $\kappa_r$ is that capability's minimum meaningful change. A trained child is promoted only when the primary target gain is at least $+0.5\pp$, every non-primary capability changes by at least $-1\pp$, $U_{\mathrm{prom}}>0$, all retention constraints pass, and no hard execution failure occurs. Otherwise the parent is restored. The final episode-level IFEval guardrail is $-2\pp$ relative to the initial parent for both prompt- and instruction-level strict accuracy. The one-point task MCID defines retrospective outcome labels, the two-point normalization scale $\kappa_i=.02$ above determines when source strength saturates, and the $+0.5\pp$ threshold here is the distinct minimum target gain for adopting a newly trained child. All authorization policies share this adoption rule.

\paragraph{Comparators.} Flat-Additive replaces the product and veto with $Z_i^{\mathrm{flat}}=(S_i+A_i+1-H_i)/3$, so a conflict is an ordinary additive feature rather than a hard rejection. Its policy-specific thresholds are selected on $D_{\mathrm{cal}}$ and frozen before the evaluated outcomes open. The Audit-24 Flat-Additive row uses this same rule. Additive+Veto retains the hard rejection but uses $Z_i^{\mathrm{add}}=(S_i+A_i)/2$ with calibrated thresholds $.45/.75$. Validate-All ignores $S_i$, $A_i$, and $H_i$ after the shared executability check and validates every executable candidate. All downstream validation, training, adoption, and accounting rules are shared.

% Standalone float module; include from the project root.
% Source label: tab:baseline-contracts
% Float kind: table
\begin{table*}[!t]
\centering
\scriptsize
\caption{Operational contracts for the shared-model methods in the endpoint table. The first four policies share seed-indexed candidate streams and adoption rule $g$; the remaining rows are descriptive baselines with the stated action spaces.}
\label{tab:baseline-contracts}
\begin{tabularx}{\textwidth}{@{}p{0.14\textwidth}p{0.12\textwidth}p{0.12\textwidth}Y p{0.20\textwidth}@{}}
\toprule
Method & Historical library & New proposals & Pre-full-training policy & Candidate stream and adoption \\
\midrule
BCIT & frozen & at most 3 & product, hard veto, selective validation & matched stream; shared $g$ \\
Flat-Additive & same as BCIT & same as BCIT & additive no-veto score; otherwise matched & matched stream; shared $g$ \\
Validate-All & same candidates & same as BCIT & validate every executable candidate & matched stream; shared $g$ \\
Additive+Veto & same as BCIT & same as BCIT & additive score with hard veto and BCIT validation & matched stream; shared $g$ \\
BCIT-Retrieve & frozen & disabled & BCIT authorization & retrieved library candidates only; shared $g$ \\
No-memory search & disabled & target-only & every new proposal requires current-parent validation & method-specific proposal stream; shared $g$ \\
Source-only & frozen & disabled & rank and authorize by $S_i$; no $A_i$, $H_i$, or validation & frozen historical stream; shared $g$ \\
Reuse-all & frozen & disabled & train every source-eligible candidate while a full lease fits & frozen order; shared $g$ \\
Static multitask SFT & none & none & fixed 1:1:1 optimizer-token mixture & fixed joint schedule; frozen feasible-checkpoint selection \\
\bottomrule
\end{tabularx}
\end{table*}

% Standalone float module; include from the project root.
% Source label: tab:cases
% Float kind: table
\begin{table*}[!t]
\centering
\small
\caption{Constructed authorization cases illustrating the frozen rule. These examples explain routing only; they are not experimental observations and do not contribute to any reported result.}
\label{tab:cases}
\begin{tabularx}{\textwidth}{@{}p{0.19\textwidth}c c c c p{0.17\textwidth}Y@{}}
\toprule
Illustrative update & $S_i$ & $A_i$ & $H_i$ & $\chi_i$ & Decision & Reason \\
\midrule
Rationale-first SQL SFT in a matched execution setting & 1.00 & .80 & 0 & 1 & \T & $Q_i=.80\ge\tau_h$ and replicated source evidence permits direct training \\
Low-rate finance continuation with single-run evidence & .45 & .90 & 0 & 0 & \V & $Q_i=.405$ is not rejected, but grade B evidence cannot bypass current-state validation \\
Function-calling update under an incompatible output schema & .90 & .80 & 1 & 1 & \R & The named protocol conflict is non-compensable even though the positive score is high \\
\bottomrule
\end{tabularx}
\end{table*}

\section{Data, Training, and Evaluation}

\subsection{Dataset Roles}

FinQA, Spider, and xLAM provide source update experience \cite{Chen2021FinQA,Yu2018Spider,Zhang2024xLAM}; TAT-QA, BIRD, and BFCL measure transfer to related target contexts \cite{Zhu2021TATQA,Li2023BIRD,BFCL2024}; IFEval supplies a common instruction-following retention guardrail \cite{Zhou2023IFEval}. Training, validation, adoption, and final-evaluation records are disjoint by role even when they originate from the same public benchmark.

% Standalone float module; include from the project root.
% Source label: tab:data
% Float kind: table
\begin{table}[t]
\centering
\small
\caption{Capability families and frozen target-evaluation cohorts. Counts are evaluation examples, not training-set sizes.}
\label{tab:data}
\begin{tabular}{@{}llll@{}}
\toprule
Family & Source & Target & Metric / count \\
\midrule
Finance & FinQA & TAT-QA & Official F1 / 1,663 \\
Text-to-SQL & Spider & BIRD & Exec. acc. / 400 \\
Function calling & xLAM & BFCL & Full match / 444 \\
\bottomrule
\end{tabular}
\end{table}

IFEval contains 541 prompts and 834 verifiable instructions. Prompt-level and instruction-level strict accuracy are reported separately. Target improvement is required; IFEval is a non-compensable feasibility condition and is not averaged with the target to make an update appear beneficial.

\subsection{Common Training Defaults}

Candidate cards change one declared primary factor and otherwise inherit Table~\ref{tab:train}. Full-run length remains candidate-specific and is recorded by the execution plan; short continuations and longer schedules are not forced to share an artificial step count.

% Standalone float module; include from the project root.
% Source label: tab:train
% Float kind: table
\begin{table}[t]
\centering
\small
\caption{Common full-parameter SFT defaults.}
\label{tab:train}
\begin{tabular}{@{}ll@{}}
\toprule
Field & Value \\
\midrule
Base model & Qwen3-4B \\
Framework & LlamaFactory / PyTorch \\
Precision & bfloat16 \\
Distributed optimizer & DeepSpeed ZeRO-2 \\
Context length & 4,096 tokens \\
Learning rate & $6\times10^{-6}$ \\
Scheduler / warmup & Cosine / .03 \\
Weight decay & .01 \\
Max. gradient norm & 1.0 \\
Per-device batch / accumulation & 4 / 2 \\
Effective batch on four GPUs & 32 \\
Prompt template & Qwen3 chat \\
\bottomrule
\end{tabular}
\end{table}

The method-owned compute charge is
\begin{equation}
C=\sum_j n_j(t_j^{\mathrm{end}}-t_j^{\mathrm{start}}),
\end{equation}
including training, validation, method-triggered evaluation, loading, and failed method jobs. Common final scoring is not used to create a policy-specific advantage.

\paragraph{Deterministic evaluation.} Endpoint generation uses one sample per prompt, temperature zero, and top-$p=1$. Finance uses the frozen official-style numerical parser, SQL uses execution against the declared database backend, BFCL uses the official structured-call checker, and IFEval uses the official strict evaluator. An API error, missing prediction, incompatible output protocol, or unusable checkpoint is a hard failure rather than a silent zero inside a mean.

\subsection{Intervention Record Example}

Table~\ref{tab:card} shows the controller-visible portion of a historical record. The target-context full outcome is deliberately excluded at authorization time.

% Standalone float module; include from the project root.
% Source label: tab:card
% Float kind: table
\begin{table*}[!t]
\centering
\small
\caption{Example frozen historical update record. This schema/provenance example is not an additional experimental outcome.}
\label{tab:card}
\begin{tabularx}{\textwidth}{@{}p{0.21\textwidth}Y@{}}
\toprule
Field & Frozen value \\
\midrule
Semantic update & Rationale-first SQL SFT \\
Atomic change & Add 260 train-only SQL rationale examples to the base-plus-retention SFT recipe \\
Required conditions & Executable SQLite databases and schema context; execution accuracy as the primary metric; compatible parent family and training stage \\
Named conflicts & Dialect/backend mismatch; incompatible output schema; missing required database access \\
Provenance & Training data, configuration, evaluator, source outcome, and run identifier each carry a content digest \\
Decision-time visibility & Source record, evidence grade, applicability fields, and provenance; no target full-run metric or held-out target error case \\
\bottomrule
\end{tabularx}
\end{table*}

\section{RQ1: Retrospective Transfer Diagnostic}

The 24 candidate--target-context pairs share seed 20260713. Each control/intervention pair uses the same parent, evaluation examples, data scale, and budget. Finance contains five target increases and three decreases; SQL contains three increases, two neutral outcomes, and three decreases; Function Calling contains three increases, one neutral outcome, and four decreases. Overall, 11/24 target effects are positive, 3/24 neutral, and 10/24 negative. Only three of the eleven target-improving updates also improve both retention measures.

Table~\ref{tab:retro} exposes two distinct failure modes. First, semantically related tasks do not guarantee positive transfer: every family contains increases and decreases. Second, target-only selection is unsafe. Standard SQL SFT improves BIRD by $2.250\pp$ while reducing IFEval-P/I by $22.736/18.705\pp$; random-sampling function SFT improves BFCL by $1.351\pp$ while reducing IFEval-P/I by $11.091/8.153\pp$.

% Standalone float module; include from the project root.
% Source label: tab:retro
% Float kind: table
\begin{table*}[!t]
\centering
\scriptsize
\caption{All Retrospective-24 matched effects in percentage points. Each row is intervention minus its matched control. P/I denote IFEval prompt/instruction strict accuracy.}
\label{tab:retro}
\begin{tabular}{@{}llrrr@{\hspace{12pt}}llrrr@{}}
\toprule
Family & Semantic update & Target $\Delta$ & P $\Delta$ & I $\Delta$ & Family & Semantic update & Target $\Delta$ & P $\Delta$ & I $\Delta$ \\
\midrule
Fin. & Balanced numeric repair & $-.487$ & $+1.109$ & $+2.038$ & SQL & Complex-query replay & $+.750$ & $.000$ & $-.120$ \\
Fin. & Strict-CoT repair & $-.890$ & $+8.133$ & $+7.914$ & SQL & Execution-reward GRPO & $+.750$ & $+1.109$ & $+.360$ \\
Fin. & CoT cold start & $+.355$ & $-.739$ & $-.719$ & SQL & Mixed execution GRPO & $-.750$ & $.000$ & $-.600$ \\
Fin. & Exact-reward GRPO & $-.182$ & $+1.479$ & $+.839$ & SQL & Hard-query GRPO & $.000$ & $+.924$ & $+.360$ \\
Fin. & Extended SFT & $+.357$ & $+.370$ & $+1.679$ & SQL & Rationale-first SFT & $-2.250$ & $-1.109$ & $-1.559$ \\
Fin. & Precision replay & $+.352$ & $+13.309$ & $+11.990$ & SQL & Rejection distillation & $-.250$ & $+2.773$ & $+2.398$ \\
Fin. & Filtered SFT & $+.018$ & $+1.294$ & $.000$ & SQL & Schema-reward GRPO & $.000$ & $+2.218$ & $+.839$ \\
Fin. & Short continuation & $+.092$ & $-.924$ & $-1.319$ & SQL & Standard SFT & $+2.250$ & $-22.736$ & $-18.705$ \\
\addlinespace
Func. & Argument repair & $+.676$ & $-.739$ & $-.600$ & Func. & Random-sampling SFT & $+1.351$ & $-11.091$ & $-8.153$ \\
Func. & Complexity stratification & $-2.703$ & $+2.403$ & $+1.079$ & Func. & Recovery GRPO & $.000$ & $-.185$ & $+.240$ \\
Func. & Format-retention replay & $-.225$ & $-1.848$ & $-.839$ & Func. & Reward-ablated GRPO & $+.450$ & $-.185$ & $-.839$ \\
Func. & Low-rate continuation & $-.225$ & $-.739$ & $-.120$ & Func. & Gold-filtered SFT & $-.225$ & $+1.664$ & $+.600$ \\
\bottomrule
\end{tabular}
\end{table*}

\section{RQ2: Outcome-Blind Authorization Audit}

After decisions freeze, a unit is Beneficial (B) when it reaches the target MCID, satisfies retention constraints, and has no hard failure; Harmful (H) when the target degrades, a retention bound is violated, or a hard failure occurs; and Neutral (N) otherwise. Audit-24 contains 10 B, 8 H, and 6 N units. Each audit action is resolved to Authorize or Non-Authorize by the policy's frozen decision rule and prespecified fallback using decision-time evidence only; the full-run label is opened only for scoring.

For policy $\pi$, let $B_\pi,H_\pi,N_\pi$ denote authorized counts. We report
\begin{align}
\mathrm{Coverage}(\pi)&=(B_\pi+H_\pi+N_\pi)/24,\\
\mathrm{HER}(\pi)&=H_\pi/8,\qquad \mathrm{BCov}(\pi)=B_\pi/10,\\
\mathrm{SelU}_2(\pi)&=(B_\pi-2H_\pi)/24.
\end{align}
Harm/Auth. is $H_\pi/(B_\pi+H_\pi+N_\pi)$. These count-based metrics define the audit comparison without imposing a post-hoc ordering over tied decisions.

% Standalone float module; include from the project root.
% Source label: tab:audit
% Float kind: table
\begin{table*}[!t]
\centering
\small
\caption{Audit-24 authorization results and structural ablations. Lower HER and Harm/Auth. are better; higher BCov is better.}
\label{tab:audit}
\begin{tabular}{@{}lcccccc@{}}
\toprule
Policy & Auth. B/H/N & Coverage & HER & BCov & Harm/Auth. & $\mathrm{SelU}_2$ \\
\midrule
Reuse all & 10/8/6 & 100.0\% & 100.0\% & 100.0\% & 33.3\% & $-.250$ \\
Source only & 9/6/4 & 79.2\% & 75.0\% & 90.0\% & 31.6\% & $-.125$ \\
Applicability only & 8/5/4 & 70.8\% & 62.5\% & 80.0\% & 29.4\% & $-.083$ \\
Flat-Additive & 8/5/3 & 66.7\% & 62.5\% & 80.0\% & 31.3\% & $-.083$ \\
\textbf{BCIT} & \textbf{9/2/3} & \textbf{58.3\%} & \textbf{25.0\%} & \textbf{90.0\%} & \textbf{14.3\%} & $\mathbf{+.208}$ \\
BCIT w/o candidate-specific applicability & 9/6/3 & 75.0\% & 75.0\% & 90.0\% & 33.3\% & $-.125$ \\
Additive+Veto & 8/3/3 & 58.3\% & 37.5\% & 80.0\% & 21.4\% & $+.083$ \\
BCIT w/o hard veto & 9/4/3 & 66.7\% & 50.0\% & 90.0\% & 25.0\% & $+.042$ \\
\bottomrule
\end{tabular}
\end{table*}

The paired BCIT--Flat-Additive discordance table is: harmful units, both 2 / Flat only 3 / BCIT only 0 / neither 3; beneficial units, 8 / 0 / 1 / 1; neutral units, 3 / 0 / 0 / 3. The exact two-sided McNemar test on harmful decisions gives $p=.25$, so the result is directional cohort evidence rather than a precise population error-rate estimate.

% Standalone float module; include from the project root.
% Source label: tab:audit-ci
% Float kind: table
\begin{table}[t]
\centering
\small
\caption{Audit-24 95\% marginal exact-binomial intervals.}
\label{tab:audit-ci}
\begin{tabular}{@{}lcc@{}}
\toprule
Metric & BCIT & Flat-Additive \\
\midrule
Coverage & $[36.6,77.9]$ & $[44.7,84.4]$ \\
HER & $[3.2,65.1]$ & $[24.5,91.5]$ \\
BCov & $[55.5,99.7]$ & $[44.4,97.5]$ \\
Harm/Auth. & $[1.8,42.8]$ & $[11.0,58.7]$ \\
\bottomrule
\end{tabular}
\end{table}

\section{RQ3: Bounded-Validation Fidelity}

All 24 frozen candidates continue to full training regardless of the short-run outcome, avoiding pass-only selection. At the primary 20\% nominal budget, short and full target directions agree for 20/24 candidates (83.3\%; two-sided exact binomial 95\% interval $[62.6\%,95.3\%]$) and Spearman $\rho=.720$. This marginal interval describes the evaluated units and is not cluster-robust. SFT and GRPO each achieve 10/12 sign agreement, with $\rho=.755$ and $.671$, respectively. Pass versus non-Pass gives TP/FP/FN/TN $=8/2/2/12$, corresponding to 80.0\% precision, 80.0\% beneficial recall, and 14.3\% false-positive rate. The four sign reversals are S04, S07, G04, and G10.

% Standalone float module; include from the project root.
% Source label: fig:fidelity
% Float kind: figure
\begin{figure}[t]
\centering
\includegraphics[width=\columnwidth]{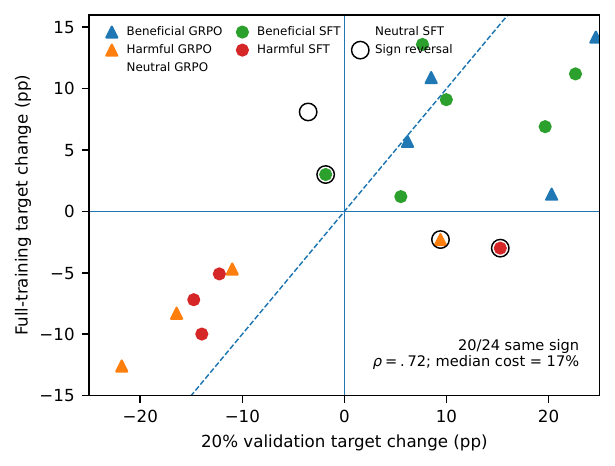}
\caption{Candidate-level bounded-validation fidelity. Circles and triangles denote SFT and GRPO; open rings mark the four sign reversals.}
\label{fig:fidelity}
\end{figure}

% Standalone float module; include from the project root.
% Source label: tab:shortfull
% Float kind: table
\begin{table*}[!t]
\centering
\scriptsize
\caption{Candidate-level ShortFull-24 ledger at the primary 20\% validation budget. Short and full target changes are raw score fractions.}
\label{tab:shortfull}
\begin{tabular}{@{}lllrrl@{\hspace{10pt}}lllrrl@{}}
\toprule
ID & Type & Label & Short $\Delta$ & Full $\Delta$ & Action & ID & Type & Label & Short $\Delta$ & Full $\Delta$ & Action \\
\midrule
S01&SFT&H&$-.13974$&$-.100$&Fail&G01&GRPO&H&$-.21813$&$-.126$&Fail\\
S02&SFT&H&$-.14760$&$-.072$&Fail&G02&GRPO&H&$-.16441$&$-.083$&Fail\\
S03&SFT&H&$-.12255$&$-.051$&Fail&G03&GRPO&H&$-.10990$&$-.047$&Fail\\
S04&SFT&H&$+.15270$&$-.030$&Pass&G04&GRPO&H&$+.09411$&$-.023$&Pass\\
S05&SFT&N&$-.06664$&$-.012$&Fail&G05&GRPO&N&$-.22667$&$-.010$&Fail\\
S06&SFT&B&$+.05525$&$+.012$&Pass&G06&GRPO&N&$-.09494$&$-.003$&Fail\\
S07&SFT&B&$-.01839$&$+.030$&Inconc.&G07&GRPO&B&$+.20300$&$+.014$&Pass\\
S08&SFT&N&$+.04449$&$+.048$&Inconc.&G08&GRPO&N&$+.01545$&$+.034$&Inconc.\\
S09&SFT&B&$+.19669$&$+.069$&Pass&G09&GRPO&B&$+.06183$&$+.057$&Pass\\
S10&SFT&B&$+.09983$&$+.091$&Pass&G10&GRPO&N&$-.03553$&$+.081$&Inconc.\\
S11&SFT&B&$+.22638$&$+.112$&Pass&G11&GRPO&B&$+.08500$&$+.109$&Pass\\
S12&SFT&B&$+.07644$&$+.136$&Pass&G12&GRPO&B&$+.24632$&$+.142$&Inconc.\\
\bottomrule
\end{tabular}
\end{table*}

% Standalone float module; include from the project root.
% Source label: tab:budget-sens
% Float kind: table
\begin{table}[t]
\centering
\small
\caption{Short-budget sensitivity on the same 24 candidates. P/F/I denote Pass/Fail/Inconclusive counts.}
\label{tab:budget-sens}
\begin{tabular}{@{}ccccc@{}}
\toprule
Budget & P/F/I & Sign agree & $\rho$ & Median cost/full \\
\midrule
5\% & 9/7/8 & 62.5\% & .42 & .06 \\
10\% & 10/8/6 & 75.0\% & .61 & .11 \\
20\% & 10/9/5 & 83.3\% & .72 & .17 \\
\bottomrule
\end{tabular}
\end{table}

The 20\% result is consistent with the primary operating point without implying that more validation is always optimal. Four sign reversals and two false Pass cases remain; a bounded-validation result provides current-parent evidence but does not replace full training or the shared adoption rule.

\section{RQ4: Matched 36-GPU-Hour Episodes}

BCIT, Flat-Additive, Validate-All, and Additive+Veto use the same six seed-indexed starts and candidate streams, a common adoption rule, and a 36-GPU-hour cap. The amendment prespecified BCIT--Validate-All as the first comparison and BCIT--Additive+Veto as the second before either corresponding baseline outcome was opened; matched pairs also share data order. The two component ablations use S1--S3 and are descriptive.

For the anytime endpoint, the frozen grid is $\{0,6,12,18,24,30,36\}$ GPU-hours. With $M(b)$ the incumbent parent checkpoint's cross-task mean,
\begin{equation}
\mathrm{AUC}=\frac{1}{36}\sum_{k=1}^{6}\frac{M(b_{k-1})+M(b_k)}{2}(b_k-b_{k-1}).
\end{equation}
If an episode stops early, its final adopted parent is carried forward.

\subsection{Primary Paired Comparison}

% Standalone float module; include from the project root.
% Source label: tab:primary-seeds
% Float kind: table
\begin{table*}[!t]
\centering
\scriptsize
\caption{Seed-level BCIT and Flat-Additive endpoints. Mean is recomputed from the three displayed target scores. Pair $\Delta$ is BCIT mean minus the matched Flat-Additive mean.}
\label{tab:primary-seeds}
\begin{tabular}{@{}llrrrrrrrr@{}}
\toprule
Method & Seed & TAT-QA & BIRD & BFCL & Mean & IFEval-P & IFEval-I & GPUh & Pair $\Delta$ \\
\midrule
BCIT&S1&35.400&44.000&60.586&46.662&76.895&83.094&34.7&$+2.559$\\
BCIT&S2&36.500&44.500&61.261&47.420&76.710&82.854&35.3&$+2.726$\\
BCIT&S3&35.800&43.750&60.811&46.787&76.710&82.974&35.0&$+2.351$\\
BCIT&S4&36.100&44.500&60.811&47.137&76.525&82.614&34.8&$+3.134$\\
BCIT&S5&35.100&44.000&60.360&46.487&76.895&83.094&35.6&$+1.834$\\
BCIT&S6&36.400&44.750&61.261&47.470&76.710&82.854&34.9&$+3.176$\\
\addlinespace
Flat-Add.&S1&33.200&41.000&58.108&44.103&75.046&81.055&35.9&\NA\\
Flat-Add.&S2&33.800&41.500&58.784&44.695&74.861&80.815&36.0&\NA\\
Flat-Add.&S3&33.500&41.250&58.559&44.436&75.231&81.175&35.8&\NA\\
Flat-Add.&S4&32.900&41.000&58.108&44.003&74.861&80.695&36.0&\NA\\
Flat-Add.&S5&33.700&41.250&59.009&44.653&75.046&81.055&35.7&\NA\\
Flat-Add.&S6&33.300&41.250&58.333&44.294&74.861&80.815&35.9&\NA\\
\bottomrule
\end{tabular}
\end{table*}

The paired cross-task mean difference is $2.630\pm.506\pp$ with 95\% paired $t$ interval $[2.099,3.161]$. All six differences are positive; the exact two-sided sign-flip test gives $2/64=.03125$. Task-level paired differences are $+2.483$ $[1.783,3.184]$ for TAT-QA, $+3.042$ $[2.621,3.462]$ for BIRD, and $+2.365$ $[1.791,2.939]$ for BFCL. IFEval-P/I paired differences are $+1.756$ $[1.594,1.919]$ and $+1.979$ $[1.874,2.084]$.

\subsection{Protocol-Amendment Baselines}

% Standalone float module; include from the project root.
% Source label: tab:p0-seeds
% Float kind: table
\begin{table*}[!t]
\centering
\scriptsize
\caption{Seed-level endpoints for Validate-All and Additive+Veto. $\Delta$ is the matched BCIT mean minus the baseline mean.}
\label{tab:p0-seeds}
\begin{tabular}{@{}llrrrrrrrrr@{}}
\toprule
Policy & Seed & TAT-QA & BIRD & BFCL & Mean & IFEval-P & IFEval-I & GPUh & AUC & $\Delta$ \\
\midrule
Validate-All&S1&34.720&42.600&59.350&45.557&75.620&81.740&35.74&44.280&$+1.105$\\
&S2&34.300&42.180&59.110&45.197&75.310&81.420&35.91&44.050&$+2.223$\\
&S3&34.880&42.740&59.440&45.687&75.800&81.910&35.62&44.370&$+1.100$\\
&S4&34.510&42.350&59.280&45.380&75.440&81.580&35.96&44.130&$+1.757$\\
&S5&35.020&42.810&59.520&45.783&75.930&82.030&35.68&44.420&$+.704$\\
&S6&34.460&42.490&59.200&45.383&75.550&81.690&35.83&44.190&$+2.087$\\
\addlinespace
Additive+Veto&S1&35.100&43.400&60.050&46.183&76.120&82.240&35.20&44.520&$+.479$\\
&S2&34.850&43.120&59.760&45.910&75.830&81.930&35.42&44.370&$+1.510$\\
&S3&35.320&43.560&60.140&46.340&76.350&82.410&35.06&44.640&$+.447$\\
&S4&34.980&43.250&59.910&46.047&75.960&82.080&35.37&44.430&$+1.090$\\
&S5&35.240&43.490&60.020&46.250&76.220&82.310&35.12&44.590&$+.237$\\
&S6&34.740&43.080&59.680&45.833&75.740&81.870&35.48&44.310&$+1.637$\\
\bottomrule
\end{tabular}
\end{table*}

BCIT exceeds Validate-All by $1.496\pm.614\pp$ with 95\% paired interval $[.852,2.140]$, and exceeds Additive+Veto by $.900\pm.596\pp$ with interval $[.275,1.525]$. Each comparison has six positive differences and exact two-sided sign-flip $p=.03125$. Against Validate-All, BCIT uses $.740$ fewer GPU-hours on average, interval $[.328,1.152]$, and improves AUC by $.702$ $[.224,1.181]$. The AUC gap to Additive+Veto is $.466$ with interval $[-.003,.935]$ and is treated as a secondary trend rather than a separate superiority result.

\subsection{Component Ablations}

% Standalone float module; include from the project root.
% Source label: tab:ablations
% Float kind: table
\begin{table*}[!t]
\centering
\scriptsize
\caption{Three-seed component ablations. $\Delta$ is the matched BCIT mean minus the ablation mean on S1--S3. These rows are descriptive and receive no significance test.}
\label{tab:ablations}
\begin{tabular}{@{}llrrrrrrrrr@{}}
\toprule
Policy & Seed & TAT-QA & BIRD & BFCL & Mean & IFEval-P & IFEval-I & GPUh & AUC & $\Delta$ \\
\midrule
BCIT w/o hard veto&S1&35.000&42.650&59.600&45.750&75.420&81.520&35.46&44.230&$+.912$\\
&S2&34.400&42.150&59.000&45.183&74.960&81.080&35.71&43.960&$+2.237$\\
&S3&34.700&42.420&59.300&45.473&75.210&81.310&35.58&44.100&$+1.314$\\
\addlinespace
BCIT-Reject-Unresolved&S1&34.700&42.500&59.000&45.400&76.220&82.300&32.84&44.030&$+1.262$\\
&S2&34.150&42.050&58.700&44.967&75.910&82.020&33.12&43.790&$+2.453$\\
&S3&34.450&42.280&58.900&45.210&76.080&82.180&32.96&43.910&$+1.577$\\
\bottomrule
\end{tabular}
\end{table*}

Removing the hard veto lowers the mean by $1.487\pm.679\pp$ relative to BCIT on S1--S3. BCIT-Reject-Unresolved, which rejects every candidate routed to validation, lowers it by $1.764\pm.617\pp$. It uses less compute because it declines unresolved candidates; this is not an equal-quality efficiency result.

\subsection{Endpoint and Process Summaries}

% Standalone float module; include from the project root.
% Source label: tab:endpoints
% Float kind: table
\begin{table*}[!t]
\centering
\scriptsize
\caption{All 36-GPU-hour endpoints. Task and retention columns are means; cross-task Mean and GPUh include sample SD. Secondary baselines are descriptive aggregates and are not used for inferential claims.}
\label{tab:endpoints}
\begin{tabular}{@{}lrrrrrrrrr@{}}
\toprule
Method & TAT-QA & BIRD & BFCL & Mean & Worst $\Delta$ & IFEval-P & IFEval-I & GPUh & AUC \\
\midrule
Base & 31.3&39.0&57.2&42.5&0.0&76.5&82.7&0.0&42.5\\
\multicolumn{10}{l}{\emph{Single-task specialists (36 GPU-hours each)}}\\
Finance specialist&39.6&\NA&\NA&\NA&\NA&\NA&\NA&36.0&\NA\\
SQL specialist&\NA&49.3&\NA&\NA&\NA&\NA&\NA&36.0&\NA\\
Function specialist&\NA&\NA&65.1&\NA&\NA&\NA&\NA&36.0&\NA\\
\multicolumn{10}{l}{\emph{Shared-model baselines}}\\
Flat-Additive&33.4&41.2&58.5&$44.4\pm0.3$&$+1.3$&75.0&80.9&$35.9\pm0.1$&43.6\\
Validate-All&34.6&42.5&59.3&$45.5\pm0.2$&$+2.1$&75.6&81.7&$35.8\pm0.1$&44.2\\
Additive+Veto&35.0&43.3&59.9&$46.1\pm0.2$&$+2.7$&76.0&82.1&$35.3\pm0.2$&44.5\\
No-memory search&33.2&40.8&57.8&$43.9\pm0.1$&$+0.6$&75.3&81.2&$35.6\pm0.1$&43.5\\
Source-score only&33.0&40.3&57.2&$43.5\pm0.2$&0.0&74.3&80.4&$35.3\pm0.1$&43.2\\
Static multitask SFT&32.3&39.6&56.1&$42.6\pm0.2$&$-1.1$&75.4&81.3&$35.4\pm0.2$&42.8\\
Reuse all&32.1&39.5&55.9&$42.5\pm0.1$&$-1.4$&73.6&79.3&$35.5\pm0.2$&42.6\\
\multicolumn{10}{l}{\emph{BCIT variants}}\\
BCIT-Retrieve&34.7&42.8&59.5&$45.7\pm0.3$&$+2.3$&76.0&82.0&$34.7\pm0.2$&44.3\\
BCIT w/o hard veto&34.7&42.4&59.3&$45.5\pm0.3$&$+2.1$&75.2&81.3&$35.6\pm0.1$&44.1\\
BCIT-Reject-Unresolved&34.4&42.3&58.9&$45.2\pm0.2$&$+1.7$&76.1&82.2&$33.0\pm0.1$&43.9\\
\textbf{BCIT}&\textbf{35.9}&\textbf{44.3}&\textbf{60.8}&$\mathbf{47.0\pm0.4}$&$\mathbf{+3.6}$&\textbf{76.7}&\textbf{82.9}&$35.1\pm0.3$&\textbf{44.9}\\
\bottomrule
\end{tabular}
\end{table*}

\paragraph{Promotion yield.}
Across the six paired episodes, BCIT promotes 46 of 62 fully trained candidates and rolls back 16, for a 74.2\% promotion yield. Flat-Additive promotes 28 of 63 and rolls back 35, for a 44.4\% yield. These counts use the shared adoption rule and are reported as process descriptors rather than independent statistical units.

The specialists are concentration references rather than shared-model baselines: each uses a separate full budget for one task, leaves the other two tasks and IFEval unevaluated, and does not produce a single shared model.

\paragraph{Earlier development pilot.} Under an earlier 24-GPU-hour protocol, BCIT, BCIT-Retrieve, and Flat-Additive all rolled back to Base with cross-task mean 42.514 and IFEval-P 76.525; Static Multitask retained a degraded child with cross-task mean 25.600 and IFEval-P 55.638. This null pilot is retained as protocol history and is not substituted for the finalized 36-hour study.

\section{Statistical Calculations and Claim Map}

All six-seed means and sample standard deviations are recomputed from the seed rows. Paired task, retention, and cross-task intervals use two-sided Student $t$ intervals with five degrees of freedom. The exact sign-flip test enumerates all $2^6=64$ sign assignments and assumes that paired-difference signs are exchangeable under the sharp null; with six pairs, its smallest attainable two-sided value is $.03125$. The six complete episodes, not checkpoints, candidates, tasks, or evaluation examples, are the sampling units for end-to-end inference.

Audit-24 marginal rate intervals are two-sided exact binomial intervals. The harmful-action comparison uses the paired $2\times2$ discordance table and an exact two-sided McNemar test. Because audit units may share capability or recipe structure, these intervals describe the evaluated cohort and are not presented as cluster-robust population intervals.

\paragraph{Claim-to-evidence map.}
Transfer heterogeneity is supported by Table~\ref{tab:retro}; authorization quality by Tables~\ref{tab:audit}--\ref{tab:audit-ci}; validation fidelity by Figure~\ref{fig:fidelity} and Tables~\ref{tab:shortfull}--\ref{tab:budget-sens}; equal-budget endpoint claims by Tables~\ref{tab:primary-seeds}--\ref{tab:endpoints}; and process behavior by the pooled promotion and rollback counts reported above. These links locate the candidate- or episode-level evidence reported for the corresponding main-paper claims; statistical calculations use unrounded values from the relevant tables and accompanying CSV files where provided.

\end{document}